\documentclass[conference]{IEEEtran}
\IEEEoverridecommandlockouts
\usepackage{cite}
\usepackage{hyperref}
\usepackage{booktabs}
\usepackage{amsmath,amssymb,amsfonts}
\usepackage{algorithmic}
\usepackage{graphicx}
\usepackage{textcomp}
\usepackage{xcolor}
\def\BibTeX{{\rm B\kern-.05em{\sc i\kern-.025em b}\kern-.08em
    T\kern-.1667em\lower.7ex\hbox{E}\kern-.125emX}}
\begin{document}

\title{Activity and Subject Detection for UCI HAR Dataset with \& without missing Sensor Data\\
}

\author{\IEEEauthorblockN{Debashish Saha}
\IEEEauthorblockA{\textit{dept. of Electrical Eng.} \\
\textit{Stony Brook University}\\
Stony Brook, USA \\
debashish.saha@stonybrook.edu}
\and
\IEEEauthorblockN{Piyush Malik}
\IEEEauthorblockA{\textit{dept. of Comp. Sci.} \\
\textit{Stony Brook University}\\
Stony Brook, USA \\
pimalik@cs.stonybrook.edu}
\and
\IEEEauthorblockN{Adrika Saha}
\IEEEauthorblockA{\textit{dept. of Comp. Eng.} \\
\textit{American International University Bangladesh}\\
Dhaka, Bangladesh \\
18-37359-1@student.aiub.edu}
}

\maketitle

\begin{abstract}
Current studies in Human Activity Recognition (HAR) primarily focus on the classification of activities through sensor data, while there is not much emphasis placed on recognizing the individuals performing these activities. This type of classification is very important for developing personalized and context-sensitive applications. Additionally, the issue of missing sensor data, which often occurs in practical situations due to hardware malfunctions, has not been explored yet. This paper seeks to fill these voids by introducing a lightweight LSTM-based model that can be used to classify both activities and subjects. The proposed model was used to classify the HAR dataset by UCI \cite{uci_har}, achieving an accuracy of 93.89\% in activity recognition (across six activities), nearing the 96.67\% benchmark, and an accuracy of 80.19\% in subject recognition (involving 30 subjects), thereby establishing a new baseline for this area of research. We then simulate the absence of sensor data to mirror real-world scenarios and incorporate imputation techniques, both with and without Principal Component Analysis (PCA), to restore incomplete datasets. We found that K-Nearest Neighbors (KNN) imputation performs the best for filling the missing sensor data without PCA because the use of PCA resulted in slightly lower accuracy. These results demonstrate how well the framework handles missing sensor data, which is a major step forward in using the Human Activity Recognition dataset for reliable classification tasks. 
\end{abstract}

\maketitle

\section{Introduction}
Human Activity Recognition (HAR) has become a significant research area within the field of ambient and context-aware computing. Because of its many uses, such as healthcare monitoring, abnormal behavior detection, human–computer interaction, and assistive technologies to improve the quality of life for senior citizens, the ability to recognize daily activities is becoming more and more important in our everyday life. HAR frameworks use information gathered from several sensors to make it easier to detect, identify, and categorize particular human movements or activities.

The two main categories of HAR are wearable sensor-based and vision-based techniques. There are many challenges in using video sequences for human activity recognition, including privacy concerns, high computational complexity, and limited pervasiveness, which makes it difficult to capture full-body movements while moving around \cite{ieee_paper}. All of these concerns make it difficult to use vision-based sensors in practical situations.

On the other hand, wearable sensors and the inertial sensors included in our everyday smart gadgets provide a more useful and discrete way to gather information about human activities. These sensors require little to no installation work and are small, easy to use, economical, and energy-efficient. For HAR tasks, gadgets like smartphones and smartwatches with sensors like compasses, gyroscopes, magnetometers, and accelerometers have become quite practical \cite{mdpi_paper}. It is still very difficult to generalize HAR models to different activities and sensor configurations. Because various people do the same action differently and because different people repeat the same activity at different times, activity signal patterns are very different. Recognizing the subject from the signal data is a difficult and complex undertaking since overlapping signal patterns between different activities make the classification procedure even more difficult.

Although deep learning-based Human Activity Recognition (HAR) has made great strides, current datasets and studies mostly focuses on activity classification rather than individual identification of the performers of those activities. This leaves a need for creating models capable of doing subject recognition, which could allow for more customized and context-aware applications. Furthermore, little study has been done on the problem of missing sensor data, a common occurrence in real-world settings because of hardware constraints, climatic conditions, or communication faults causing sensor failures. Missing data greatly affects the accuracy and dependability of the models, hence there is a need for managing insufficient sensor data. This work advances HAR research by suggesting a strong and scalable LSTM-based method able to provide dependable performance for both activity and subject recognition, even under missing sensor data.

\section{Related Work}

HAR datasets have seen significant progress, especially in the area of deep learning-based activity classification. For example, Ronao and Cho's study~\cite{ordonez2016} showed that Deep Convolutional and LSTM Recurrent Neural Networks may be used to recognize activities, with a \textbf{95.75\%} accuracy rate on the UCI HAR dataset. A model based on Collaborative Edge Learning (CELearning) was also presented by Xu, Tang, Jin and Pan ~\cite{Ariza2022}, and it attained an even better accuracy of \textbf{96.67\%}. These techniques demonstrate the effectiveness of multilayered deep learning models.

On the other hand, there is still much to learn about the field of subject recognition with HAR datasets. As of right now, there isn't a commonly used model or benchmark for subject classification utilizing the UCI HAR dataset. This disparity highlights the need for additional study in this field, especially for applications that call for context-aware algorithms and individualized insights.

Furthermore, one of the most important challenges for enhancing the resilience and dependability of HAR systems in practical contexts is dealing with missing sensor data. A unique method for improving activity detection without specifically recovering lost data was proposed by Hossain et al.~\cite{hossain2020}. Their approach boosted recognition accuracy by learning to handle partial datasets efficiently through the introduction of unpredictable missing data patterns throughout the training process. In a similar vein, Xaviar et al.~\cite{xaviar2023} created the Centaur multimodal fusion model, which used self-attention mechanisms to capture cross-sensor correlations and a denoising autoencoder in conjunction with convolutional layers to clean noisy data. The Centaur model maintained strong performance in HAR tasks despite noise and repeated missing data across various sensor modalities.

These experiments demonstrate some progress in handling incomplete data and activity recognition. Approaches that can handle both activity and subject detection are still  needed, particularly in situations where data is missing. To address these issues we have created a lightweight LSTM-based system that can used to classify either subjects or activities; and incorporate strong imputation techniques for handling imperfect sensor data.

\section{Methodology}

Our proposed framework uses the UCI Human Activity Recognition (HAR) dataset, which consists of data from 30 participants performing six different activities: \textit{WALKING, WALKING\_UPSTAIRS, WALKING\_DOWNSTAIRS, SITTING, LAYING, STANDING}. Our goal is to create a model that can be used to train a classifier either for activity or the subject, even in the presence of missing sensor data. 

The framework integrates three main components:
\begin{itemize}
    \item Establish baseline accuracy with and without missing data using Long Short-Term Memory (LSTM) networks.
    \item Imputation techniques to handle missing sensor data, utilizing either a Univariate Imputer (\texttt{SimpleImputer}) or a K-Nearest Neighbors Imputer (\texttt{KNNImputer}).
    \item Dimensionality reduction using Principal Component Analysis (PCA) to simplify the data while retaining essential information.
\end{itemize}

\subsection{LSTM Framework}

\begin{figure}[h]
    \centering
    \includegraphics[width=0.52\textwidth]{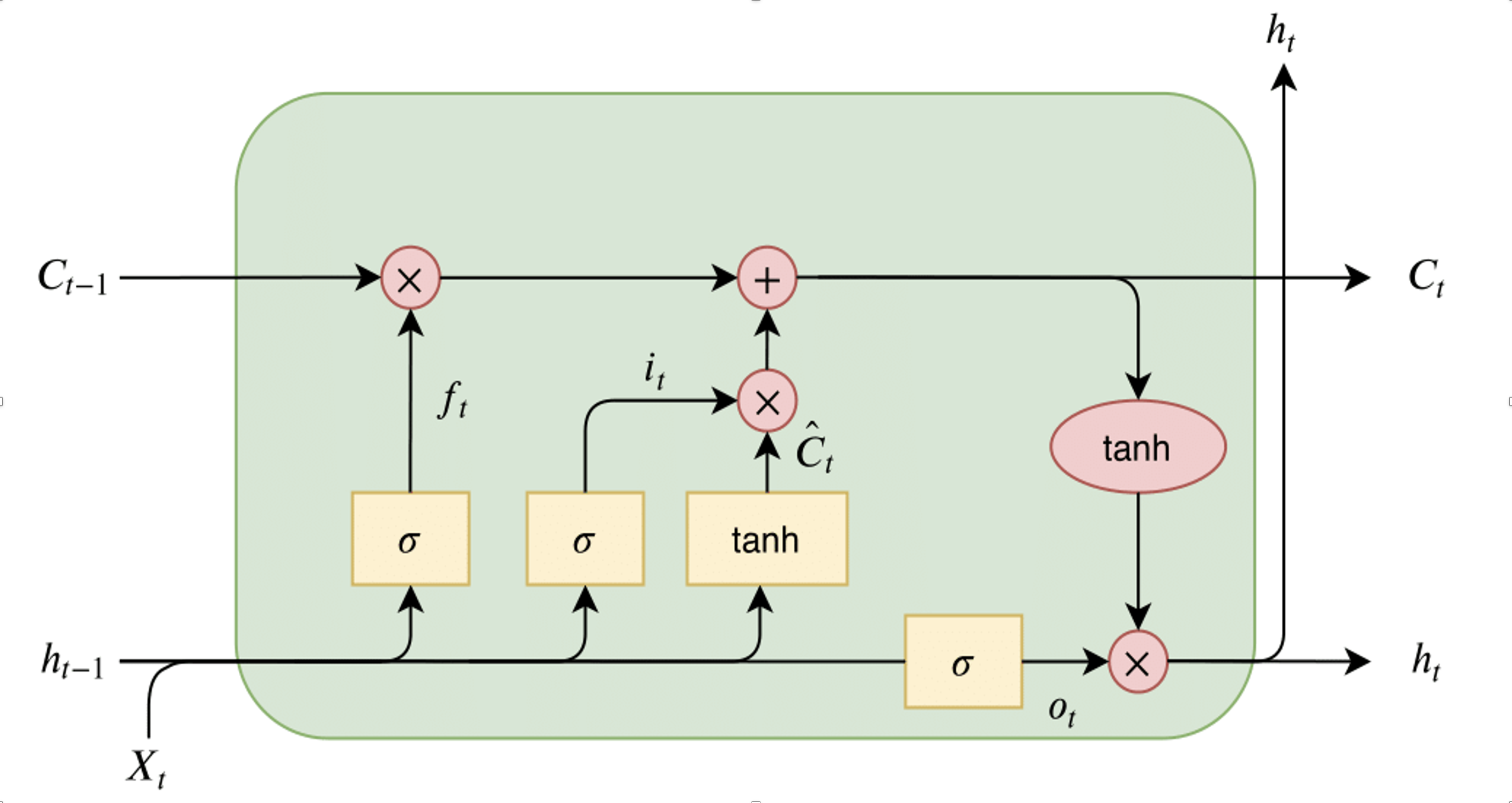} 
    \caption{LSTM Network Architecture}
    \label{fig:lstm_architecture}
\end{figure}

We construct a simple Long Short-Term Memory (LSTM) network, as shown in Fig.~\ref{fig:lstm_architecture}, to serve as the baseline model. The network is defined as follows:

Let the input sensor data sequence be represented as \( X = [x_1, x_2, \dots, x_T] \), where \( x_t \in \mathbb{R}^d \) is the sensor data at time \( t \), with \( d \) features. The Long Short-Term Memory (LSTM) processes this sequence by iteratively updating the hidden state \( h_t \) and cell state \( c_t \) at each time step using the following equations:

1. **Forget Gate**:
   \[
   f_t = \sigma(W_f x_t + U_f h_{t-1} + b_f)
   \]

2. **Input Gate**:
   \[
   i_t = \sigma(W_i x_t + U_i h_{t-1} + b_i)
   \]

3. **Output Gate**:
   \[
   o_t = \sigma(W_o x_t + U_o h_{t-1} + b_o)
   \]

4. **Candidate Cell State**:
   \[
   \tilde{c}_t = \tanh(W_c x_t + U_c h_{t-1} + b_c)
   \]

5. **Cell State Update**:
   \[
   c_t = f_t \odot c_{t-1} + i_t \odot \tilde{c}_t
   \]

6. **Hidden State Update**:
   \[
   h_t = o_t \odot \tanh(c_t)
   \]

Here, \( \sigma \) is the sigmoid activation function, \( \tanh \) is the hyperbolic tangent activation, and \( \odot \) is the element-wise multiplication operation.

\textbf{Dropout Layer
}
To prevent overfitting, a dropout layer is applied to the output of the LSTM layer. This layer randomly sets a fraction \( p \) of the hidden weights to zero during training, making sure that the model does not become overly trained on the training data. Mathematically, the dropout operation for the LSTM's final output can be expressed as:

\[
h_t^{\text{dropout}} = h_t \odot m, \quad m \sim \text{Bernoulli}(1 - p)
\]

where \( m \) is a mask vector sampled from a Bernoulli distribution with probability \( 1 - p \), and \( p \) is the dropout rate.

\textbf{Final Classification Layer
}
After processing the entire sequence, the hidden state at the final time step, \( h_T^{\text{dropout}} \), is passed through a fully connected layer followed by the softmax function to produce the predicted class probabilities:

\[
y = \text{softmax}(W_y h_T^{\text{dropout}} + b_y)
\]

where \( y \in \mathbb{R}^K \) represents the probability distribution over \( K \) activity classes.

\textbf{Loss Function
}
The model is trained by minimizing the categorical cross-entropy loss, defined as:

\[
L = -\frac{1}{N} \sum_{i=1}^N \sum_{k=1}^K y_{i,k} \log(\hat{y}_{i,k})
\]

Here:
- \( N \) is the number of samples in the training batch,
- \( K \) is the number of classes,
- \( y_{i,k} \) is the true label (1 if sample \( i \) belongs to class \( k \), and 0 otherwise),
- \( \hat{y}_{i,k} \) is the predicted probability for class \( k \) of sample \( i \).

This loss function ensures that the model learns to maximize the predicted probabilities for the correct activity or subject class.

\subsection{Handling Missing Sensor Data}

To simulate real-world scenarios, missing data is introduced randomly in the dataset. The simulations introduced missing data for durations ranging from 1 to 10 seconds intervals assuming that the accelerometer and/or the gyroscope becomes non-functional. The missing data is imputed using the following methods:

\textbf{Simple Imputation
}
Missing values \(x_{ij}\) are replaced with the mean \(\mu_j\) or median of the \(j^{th}\) feature:
\[
x_{ij} = \mu_j = \frac{1}{N} \sum x_{ij} \quad \text{(if missing)} \tag{9}
\]

\textbf{KNN Imputation
}
Missing values are estimated using the \(k\)-nearest neighbors, where the imputed value is:
\[
x_{ij} = \frac{1}{k} \sum_{k \in \text{neighbors}} x_{ik} \tag{10}
\]

\subsection{Dimensionality Reduction with PCA}

To reduce data dimensionality and noise, Principal Component Analysis (PCA) is applied. Given a dataset \(X \in \mathbb{R}^{N \times d}\), PCA projects \(X\) onto \(r\)-dimensional space (\(r < d\)):
\[
Z = XW \tag{11}
\]
Here, \(W\) contains the eigenvectors of the covariance matrix of \(X\), corresponding to the top \(r\) eigenvalues.

\section{Dataset and Experiments}

For our experiment, we used the Human Activity Recognition Using Smartphones dataset by UCI~\cite{uci_har}, which includes data collected from 30 participants performing six activities: \textit{WALKING, WALKING\_UPSTAIRS, WALKING\_DOWNSTAIRS, SITTING, LAYING, and STANDING}. The dataset comprises 10,299 entries and 561 features, divided into predefined training and testing sets. All the features were collected using only two sensors: the accelerometer and the gyroscope. Input features were normalized to ensure improved convergence and stability during model training.

The training was performed using an LSTM-based deep learning model implemented in Python. We used the Keras library with a TensorFlow backend. We designed a model architecture to balance simplicity and performance. The architecture consists of:
\begin{itemize}
    \item An LSTM layer with 128 units, which processes the input sequences from the dataset.
    \item A dropout layer with a rate of 20\%, to reduce overfitting by randomly deactivating neurons during training.
    \item Two dense layers:
    \begin{itemize}
        \item The first dense layer has 64 units with ReLU activation to capture hierarchical feature representations.
        \item The second dense layer applies a softmax activation function for multi-class classification.
    \end{itemize}
\end{itemize}
The model was compiled using the categorical crossentropy loss function and the Adam optimizer, with accuracy used as the evaluation metric.

For activity recognition, the training was straightforward. We trained the LSTM model with the training data using a 20\% validation split. For subject recognition, we had to modify the dataset. In the original dataset, there are a total of 30 subjects, but the training data includes only 21 subjects, and the test data includes 9 subjects. This distribution is not ideal for performing subject recognition. We needed a dataset where all 30 subjects were present in both the train and test sets. To achieve this, we combined the existing train and test sets into a single dataset and then used the \texttt{train\_test\_split} function provided by the \texttt{sklearn} library to split the data. Similar to the original training set, 20\% of the data was reserved for our newly created test set.

We simulated scenarios with missing sensor data to evaluate the robustness of the proposed framework under real-world conditions where sensor failures are likely. The dataset, sampled at 50 Hz, contains 345 accelerometer features and 213 gyroscope features among its 561 total features. To mimic sensor failures, we introduced missing data for durations ranging from 1 to 10 seconds, simulating instances where either the accelerometer or gyroscope becomes non-functional. We envisioned 6 such scenarios, and in each scenario, a total of 10 seconds of data was missing, which meant that approximately 24.27\% of rows in the dataset contained some missing sensor values. 

To address these missing values, we applied \texttt{SimpleImputer} and \texttt{KNNImputer} techniques, both with and without Principal Component Analysis (PCA). PCA was applied for dimensionality reduction, with 175 components retained. This number was determined based on our analysis, which showed that the retention of 175 components was necessary to preserve 99\% of the variance in the dataset.

\section{Evaluation}

We conducted training and testing for both activity and subject recognition. We trained the model for 300 epochs, and used the \texttt{Accuracy} score of the test set for our metrics.

The LSTM-based model was evaluated on the HAR dataset without missing sensor data for activity recognition. It achieved a test accuracy of \textbf{93.89\%} for classifying six activities: \textit{WALKING, WALKING\_UPSTAIRS, WALKING\_DOWNSTAIRS, SITTING, LAYING, STANDING}. It is very close to the benchmark accuracy of 96.67\% achieved by more complex Collaborative Edge Learning (CELearning) models~\cite{Ariza2022}. 

\begin{figure}[h]
    \centering
    \includegraphics[width=0.5\textwidth]{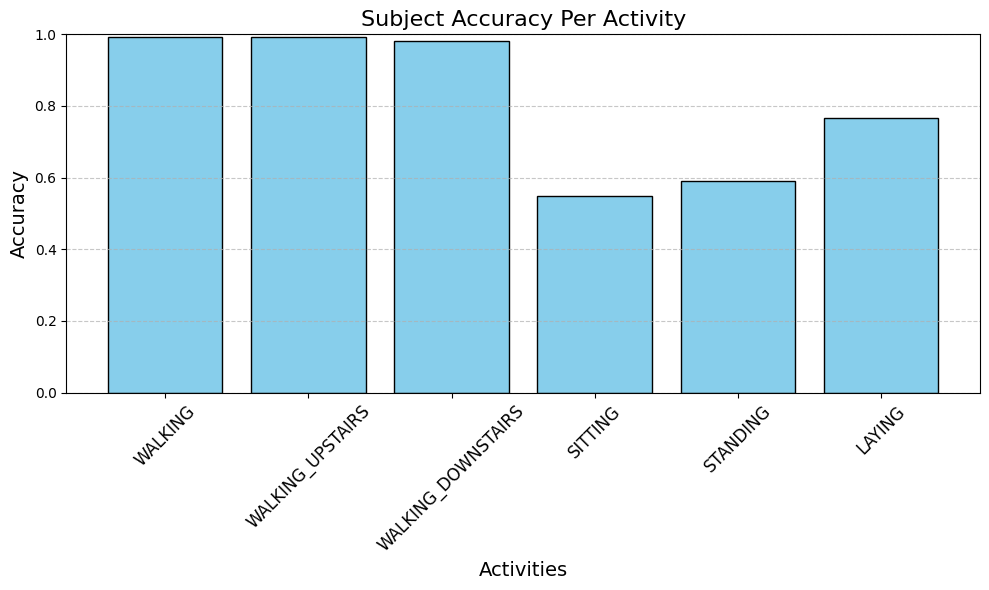} 
    \caption{Subject recognition accuracy per activity}
    \label{fig:subject_accuracy}
\end{figure}

For subject recognition, our model achieved a test accuracy of \textbf{80.19\%} for classifying 30 subjects. Currently there is no benchmark for subject recognition for this dataset. If we look at the subject class accuracy per activity, the picture becomes much clearer. Among all subjects, it is the \textit{SITTING, LAYING, STANDING} activities that cause problems. Since these are stationary activities, they might appear very similar across all subjects, as shown in Figure~\ref{fig:subject_accuracy}.

The results demonstrate the efficiency of our architecture, which uses an LSTM layer, dropout for regularization, and dense layers for classification.

\subsection{Insights from Missing Data Experiments}

The following six scenarios were created by us to evaluate the impact of missing sensor data on the proposed framework:

\begin{enumerate}
    \item \textbf{1 Second ACC \& 1 Second Gyro (5 intervals):}
    \begin{itemize}
        \item A total of 5 intervals, each lasting 1 second, where both accelerometer (ACC) and gyroscope (Gyro) data are missing.
        \item Proportion of missing data: \textbf{12.07\%}.
    \end{itemize}

    \item \textbf{5 Second ACC \& 5 Second Gyro (1 interval):}
    \begin{itemize}
        \item A single 5-second interval where both accelerometer and gyroscope data are missing.
        \item Proportion of missing data: \textbf{12.06\%}.
    \end{itemize}

    \item \textbf{5 Second ACC (2 intervals):}
    \begin{itemize}
        \item Two intervals, each lasting 5 seconds, where only accelerometer (ACC) data is missing.
        \item Proportion of missing data: \textbf{14.91\%}.
    \end{itemize}

    \item \textbf{5 Second Gyro (2 intervals):}
    \begin{itemize}
        \item Two intervals, each lasting 5 seconds, where only gyroscope (Gyro) data is missing.
        \item Proportion of missing data: \textbf{9.22\%}.
    \end{itemize}

    \item \textbf{10 Seconds ACC (1 interval):}
    \begin{itemize}
        \item A single 10-second interval where only accelerometer (ACC) data is missing.
        \item Proportion of missing data: \textbf{14.91\%}.
    \end{itemize}

    \item \textbf{10 Seconds Gyro (1 interval):}
    \begin{itemize}
        \item A single 10-second interval where only gyroscope (Gyro) data is missing.
        \item Proportion of missing data: \textbf{9.22\%}.
    \end{itemize}
\end{enumerate}

\section{Results}

The results of the missing data experiments are promising. The baseline \textbf{activity recognition} accuracy of the model was \textbf{93.85\%} under ideal conditions with no missing data. When we introduced missing sensor data, accuracy of the model reduced significantly. With the lowest accuracy dropping to \textbf{77.29\%} depending on the scenario. For \textbf{subject recognition} the accuracy was \textbf{80.19\%} with no missing data and with missing data it goes as low as \textbf{61.35\%}. Applying imputation techniques showed substantial recovery in performance as shown in Table~\ref{tab:activity_recognition} and Table~\ref{tab:subject_recognition}.

\begin{table}[h!]
\centering
\caption{Accuracy with simulated missing Data and Imputation for Activity Recognition}
\resizebox{0.53\textwidth}{!}{%
\begin{tabular}{lcccccc}
\toprule
\hline
\textbf{Missing data} & \textbf{Accuracy with} & \textbf{} & \textbf{} & \textbf{} & \textbf{} \\
\textbf{intervals} & \textbf{missing data} & \textbf{SI} & \textbf{KNN} & \textbf{SI+PCA} & \textbf{KNN+PCA} \\
\hline
\midrule
1s ACC 1s Gyro   & 0.7822 & 0.8989 & 0.8860 & 0.8975 & 0.8856 \\
(5 intervals) \\
5s ACC 5s Gyro   & 0.7872 & 0.8901 & 0.8931 & 0.8907 & 0.8928 \\
(1 interval) \\
5 Sec ACC                   & 0.7716 & 0.8663 & 0.9233 & 0.8653 & 0.9237 \\
(2 intervals) \\
5s Gyro                   & 0.7974 & 0.9189 & 0.9301 & 0.9186 & 0.9284 \\
(2 intervals) \\
10s ACC                   & 0.7710 & 0.8626 & 0.9138 & 0.8612 & 0.9138 \\
(1 intervals) \\
10s Gyro                   & 0.7740 & 0.9308 & 0.9362 & 0.9298 & 0.9348 \\
(1 interval) \\
\hline
\bottomrule
\end{tabular}%
\label{tab:activity_recognition}

}
\end{table}

\begin{table}[h!]
\centering
\caption{Accuracy with Simulated Missing Data and Imputation for Subject Recognition}
\resizebox{0.53\textwidth}{!}{%
\begin{tabular}{lccccc}
\toprule
\hline
\textbf{Missing data} & \textbf{Accuracy with} & \textbf{} & \textbf{} & \textbf{} & \textbf{} \\
\textbf{intervals} & \textbf{missing data} & \textbf{SI} & \textbf{KNN} & \textbf{SI+PCA} & \textbf{KNN+PCA} \\
\hline
\midrule
1s ACC \& 1s Gyro   & 0.6150 & 0.6417 & 0.7340 & 0.6282 & 0.7199 \\
(5 intervals) \\
5s ACC \& 5s Gyro    & 0.6184 & 0.6383 & 0.7340 & 0.6262 & 0.7170 \\
(1 interval) \\
5s ACC               & 0.6180 & 0.6413 & 0.7442 & 0.6277 & 0.7282 \\
(2 intervals) \\
5s Gyro              & 0.6141 & 0.6510 & 0.7515 & 0.6403 & 0.7398 \\
(2 intervals) \\
10s ACC              & 0.6155 & 0.6340 & 0.7476 & 0.6233 & 0.7340 \\
(1 interval) \\
10s Gyro             & 0.6170 & 0.6490 & 0.7587 & 0.6379 & 0.7481 \\
(1 interval) \\
\hline
\bottomrule
\end{tabular}%
}
\label{tab:subject_recognition}
\end{table}

From the data above we gathered couple of key insights

\begin{itemize}
    \item \textbf{Simple Imputer:} Achieved accuracy recovery across all scenarios, ranging from \textbf{86.26\%} to \textbf{93.08\%} for activity recognition and \textbf{63.40\%} to \textbf{64.90\%} for subject recognition.
    \item \textbf{KNN Imputer:} Consistently outperformed or got very close to Simple Imputer in most scenarios, achieving accuracy ranging from \textbf{88.60\%} to \textbf{93.62\%} for activity recognition (without missing data accuracy was \textbf{93.89\%}) and \textbf{73.40\%} to \textbf{75.87\%} for subject recognition (without missing data accuracy was \textbf{80.19\%}).
\end{itemize}

Combining PCA with imputation provided mixed results. While PCA reduced dimensionality, it did not consistently improve accuracy. In the majority of cases, accuracy is slightly worse. For example:
\begin{itemize}
    \item \textbf{Simple Imputer with PCA:} Simple Imputer with PCA has an accuracy score ranging from \textbf{86.12\%} to \textbf{92.98\%} for activity recognition. For subject recognition, it ranges from \textbf{62.33\%} to \textbf{64.03\%}. 
    \item \textbf{KNN Imputer with PCA:} KNN Imputer with PCA has an accuracy score ranging from \textbf{88.56\%} to \textbf{93.48\%} for activity recognition. For subject recognition, it ranges from \textbf{71.99\%} to \textbf{74.81\%}. 
\end{itemize}

\begin{figure}[h!]
    \centering
    \includegraphics[width=0.52\textwidth]{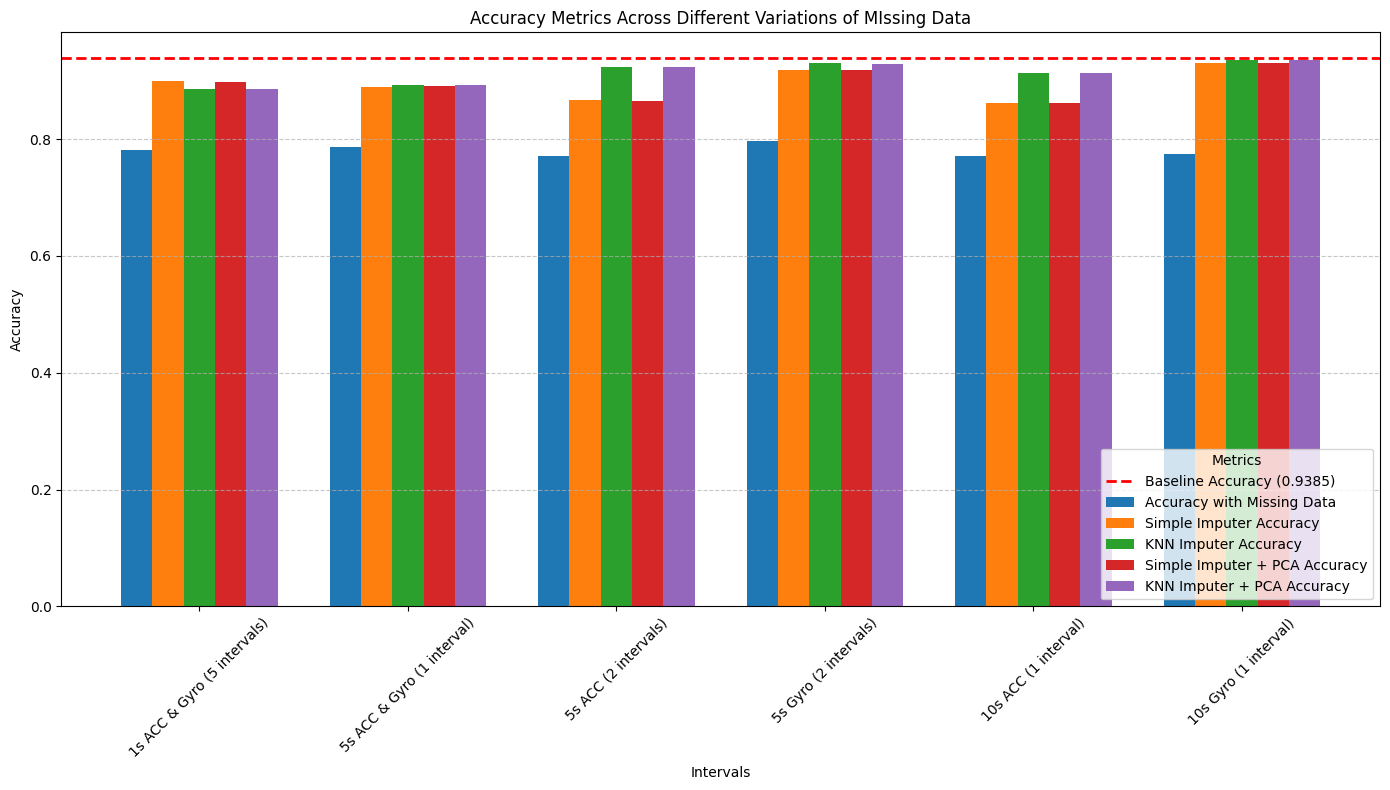} 
    \caption{Accuracy for \textbf{activity recognition} with Imputations. The blue bars represent baseline results with missing data, and the red dotted line indicates results without missing data.}
    \label{fig:activity_recognition_bar}
\end{figure}

\begin{figure}[h!]
    \centering
    \includegraphics[width=0.52\textwidth]{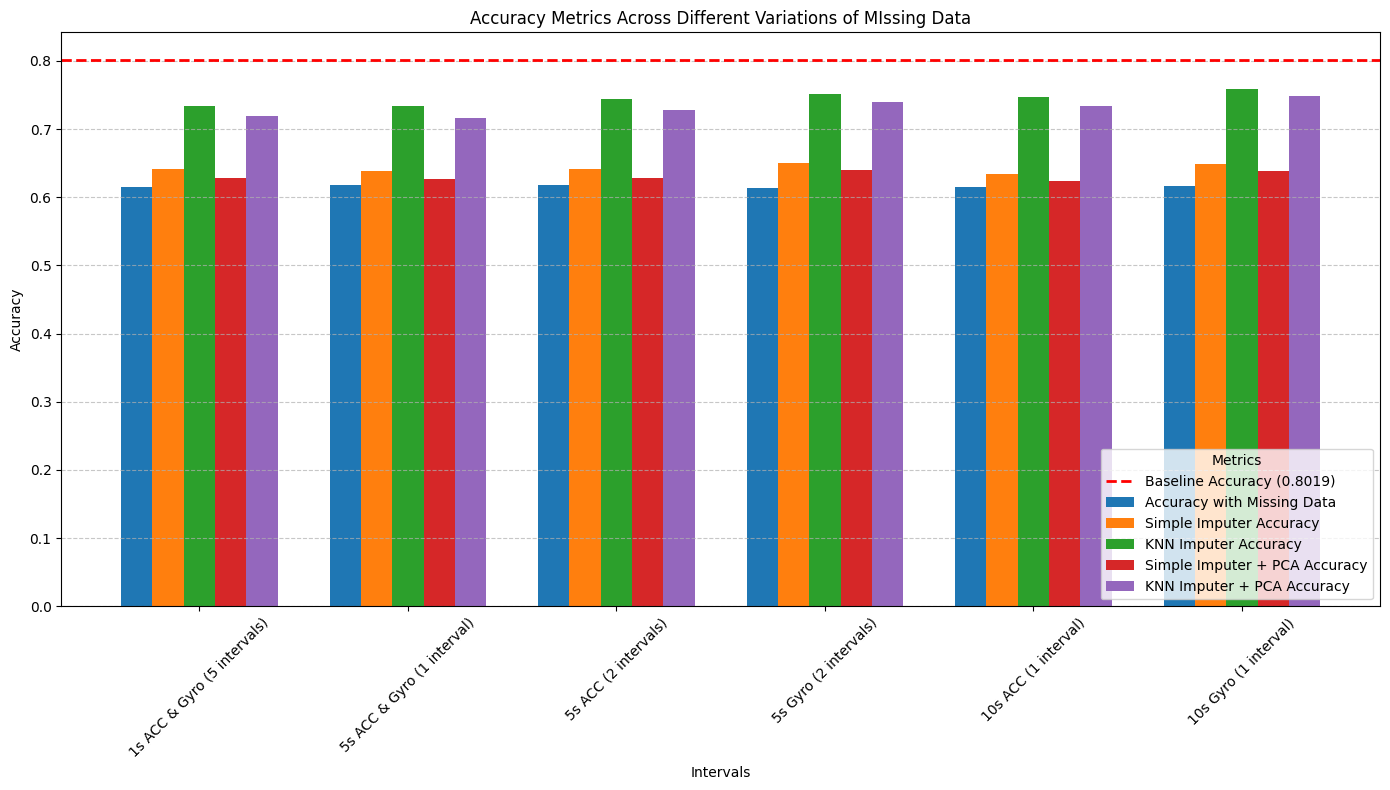} 
    \caption{Accuracy for \textbf{subject recognition} with Imputations. The blue bars represent baseline results with missing data, and the red dotted line indicates results without missing data.}
    \label{fig:subject_recognition_bar}
\end{figure}

The data also revealed that missing gyroscope data is easier to recover compared to accelerometer data. If we do a bar graph with all the accuracy metrics, the overall picture becomes much clearer. Below, you can see Figure~\ref{fig:activity_recognition_bar} for activity recognition and Figure~\ref{fig:subject_recognition_bar} for subject recognition. 

In the bar graphs:
\begin{itemize}
    \item \textbf{Red Dotted Line}: Baseline Accuracy (without missing data, ideal case).
    \item \textbf{Blue Bar}: Accuracy with Missing Data (no imputation applied).
    \item \textbf{Orange Bar}: Accuracy after applying \textbf{Simple Imputer}.
    \item \textbf{Green Bar}: Accuracy after applying \textbf{KNN Imputer}.
    \item \textbf{Red Bar}: Accuracy after applying \textbf{Simple Imputer + PCA}.
    \item \textbf{Purple Bar}: Accuracy after applying \textbf{KNN Imputer + PCA}.
\end{itemize}

As you can see all imputation techniques perform better than just with missing data and gets very close to the accuracy without missing data. These visualizations shows the effectiveness of imputation techniques and the recovery of accuracy for both activity and subject recognition tasks. This highlights the remarkable effectiveness of the KNN Imputer in handling missing sensor data independently without PCA. Our framework's ability to recover accuracy demonstrates its potential for real-world applications where sensor failures are very common.

\section{Conclusion}
Our study advances the field of Human Activity Recognition by addressing two critical gaps: identifying subjects with HAR sensor data, and the ability to handle missing sensor data. By leveraging an lightweight LSTM-based model and simulating missing data, something that happens in real world quiet often, we demonstrated the robustness and adaptability of our framework. We have established a new benchmark for subject recognition for HAR dataset and we showed that both activity and subject recognition can be easily done even with incomplete sensor data, using imputation techniques like KNN Imputer. Our work provides a foundation for creating more personalized and reliable HAR applications. Future work may explore extending our framework and try more advanced imputation methods to create a more robust classifier. The source code for this framework is publicly available at \url{https://github.com/debashish-saha1/HAR_UCI_DATASET_MISSING_DATA}.

\end{document}